\documentclass[twocolumn,10pt,letterpaper]{article}

\usepackage[margin=0.75in,columnsep=0.28in]{geometry}
\usepackage{times}
\usepackage{helvet}
\usepackage{courier}
\usepackage[hyphens]{url}
\usepackage{graphicx}
\usepackage{booktabs}
\usepackage{caption}
\usepackage{amsmath}
\usepackage{amssymb}
\usepackage{natbib}
\usepackage[colorlinks=true,linkcolor=blue,citecolor=blue,urlcolor=blue,
            breaklinks=true]{hyperref}

\newcommand{\PaperTitle}{Competence, Not Accuracy: A Diagnostic for\\
Reference-Free Judge Gates in Skill Optimization}

\begin{document}

\twocolumn[%
  \begin{center}
    {\LARGE\bfseries \PaperTitle \par}
    \vspace{1.2em}
    {\large
      Chenle Chen\textsuperscript{1} \quad
      Yangbo Wei\textsuperscript{2} \quad
      Chao Yao\textsuperscript{3} \quad
      Shaoqiang Lu\textsuperscript{2} \\[0.45em]
      Junhong Qian\textsuperscript{2} \quad
      Chen Wu\textsuperscript{4} \quad
      Lei He\textsuperscript{4}
      \par}
    \vspace{0.8em}
    {\normalsize
      \textsuperscript{1}University of California, Los Angeles \quad
      \textsuperscript{2}Shanghai Jiao Tong University \\[0.35em]
      \textsuperscript{3}Arizona State University \quad
      \textsuperscript{4}Eastern Institute of Technology, Ningbo \\[0.45em]
      \texttt{chenle@ucla.edu}
      \par}
  \end{center}
  \vspace{1.6em}
  \begin{center}
  \begin{minipage}{0.86\textwidth}
    \begin{center}\textbf{\large Abstract}\end{center}
    \vspace{0.3em}
    \noindent
Text-space skill optimization adapts a frozen agent by evolving a
natural-language skill document, accepting each candidate through a
validation gate. Existing gates rely on \emph{verifiable} rewards,
confining these methods to tasks with an automatic verifier. Replacing
the verifier with an \emph{LLM-judge gate} would lift that restriction,
but whether such a gate carries usable signal is untested. We ask a prior
question: can we tell, before placing a judge in the loop, whether its
scores separate correct from incorrect answers at all? We formalize a
reference-free judge as a \emph{latent solver}---its verdict rests on
agreement with whatever it would itself conclude, so its capacity to
evaluate is bounded by its capacity to solve. The model yields a
closed-form bound on discriminability (ROC-AUC) in the judge's competence
$c$ and answer-space size $k$, a necessary condition $c > 1/k$, and the
result that the marginal AUC is confounded by item difficulty while a
within-question estimator is not. A \emph{non-intervening} probe records
judge scores on genuine optimization runs without altering any decision.
We find discriminability at chance where competence sits near the floor
and usable above it; that a judge's benchmark accuracy overstates the
competence that matters; and, in a closed-loop study, that the screen
predicts which \emph{kind} of gating error occurs. The result is a cheap
pre-deployment diagnostic for judge gates.
  \end{minipage}
  \end{center}
  \vspace{2em}
]

\section{Introduction}
\begin{figure}[t]
\centering
\includegraphics[width=\columnwidth]{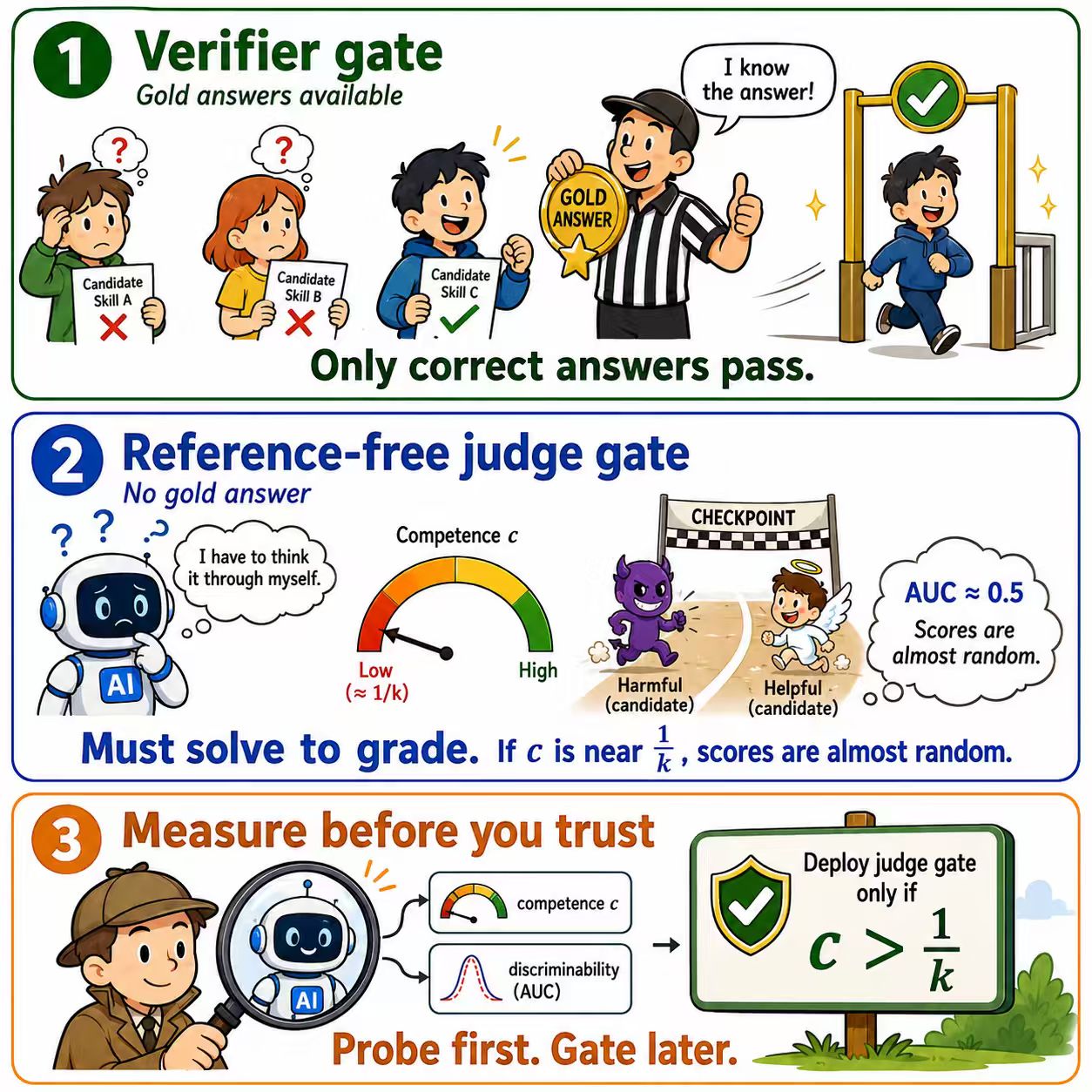}
\caption{(1)~A verifier gate has gold answers, so only correct
candidates pass. (2)~A reference-free judge has none and must in effect
re-solve each item to grade it, so once its competence $c$ nears the
chance floor $1/k$ its scores are near-random. (3)~We therefore measure
competence and discriminability \emph{before} deployment, screening out
judges that do not clear the floor---necessary, though not sufficient.}
\label{fig:motivation}
\vspace{-10pt}
\end{figure}

Recent work adapts frozen language-model agents to new domains by
evolving a compact, natural-language \emph{skill} document rather than
updating weights \citep{yang2026skilloptexecutivestrategyselfevolving,alzubi2026evoskillautomatedskilldiscovery}. These optimizers run a
candidate skill, score it, and use a \emph{validation gate} to keep the
candidate only if it improves a held-out score---directly analogous to
model selection in supervised learning. Crucially, the gate score comes
from a \emph{verifiable} signal: exact match against a gold answer, or an
executable check. This works well when such a verifier exists, but it
confines the paradigm to verifiable tasks. Figure~\ref{fig:motivation} summarizes the problem and our response.
The long-run motivation for lifting that constraint is open-ended generation---writing, dialogue,
design rationale---but we state at the outset that the present study does
not reach that far. Our model and measurements both require a single gold
answer and a verifier label, so the tasks studied here are objectively
checkable ones; extending the diagnostic to genuinely open-ended tasks,
where correctness is not binary, the answer space has no natural size,
and judging is rubric-driven rather than agreement-driven, is future
work. What we can do now is answer the prerequisite question on tasks
where ground truth is available to audit against.

A natural way to lift this restriction is to replace the verifier gate
with an \emph{LLM-judge gate}: score each candidate skill by how a judge
model rates its outputs, and accept on the judge's preference. In
reinforcement learning, judge-based signals have extended optimization
beyond verifiable rewards \citep{corbitt2025ruler}, but always in \emph{weight
space}; they have not been brought into text-space skill optimization,
where weights are frozen and the optimized object is a persistent
document. Moreover, judge reliability cannot be assumed: judges are known
to be biased and gameable, and coupling a judge to an acceptance loop
creates selection pressure that can amplify reward hacking
\citep{zheng2023judging}.

This paper asks a question that must be answered before any judge gate is
deployed: \textbf{when can a judge gate substitute for a verifier gate,
and when does it fail?} Rather than swap the gate and hope, we take a
\emph{diagnostic} approach. We instrument an existing optimizer's gate
with a judge probe that scores every candidate but \emph{does not affect
any accept/reject decision}, leaving the optimization dynamics unchanged.
This lets us measure, on real runs, whether the judge signal is
trustworthy---operationalized as \emph{discriminability}: can the judge
score separate answers the verifier marks correct from those it marks
wrong (ROC-AUC)?

Our findings are threefold. \textbf{(i)}~Judge discriminability varies
sharply across tasks---at chance on research mathematics, usable on
factual QA and graduate science---so judge signal cannot be assumed.
\textbf{(ii)}~Where the judge's own competence sits near the chance
floor, discriminability collapses, as the model predicts; and the
closed-form bound holds wherever it can be tested without difficulty
confounding. \textbf{(iii)}~A judge's headline benchmark accuracy is an
optimistic input to this test: decomposing the gap to genuine competence
shows censoring convention, benchmark exposure, and the solve/grade
context gap each contribute, all in the same direction.

\paragraph{Contributions.}
\textbf{(1)~Theory.} We model a reference-free judge as a latent solver
and derive a closed-form bound on discriminability in its competence $c$
and answer-space size $k$, a \emph{necessary but not sufficient}
condition $c > 1/k$ for any discrimination, and an identification result:
the marginal AUC is confounded by item difficulty---a judge reading only
difficulty and never the answer can score above chance---while a
within-question estimator is invariant to it.
\textbf{(2)~Measurement.} We give a non-intervening probe that records
judge signal on genuine optimization runs without altering acceptance,
together with a reusable protocol---episode-level scoring, answer-content
feeding, within-question stratification with a question-clustered
bootstrap, and proxy-based decontamination---for auditing a candidate
judge gate before deployment.
\textbf{(3)~Findings.} Discriminability collapses to chance where
competence approaches the floor, and the predicted bound holds wherever
it can be cleanly tested; a judge's headline benchmark accuracy
overstates competence, for three separable reasons that
we decompose using estimators with orthogonal error sources; and in a
closed-loop study the screen predicts which \emph{kind} of gating error
occurs---judges that fail it import regressions, while one that passes
imports none and errs only by over-rejection.

\section{Related Work}
\paragraph{Self-evolving agent skills.}
A rapidly growing line of work equips frozen agents with reusable,
natural-language skills \citep{zhang2026agentskills} and optimizes them
automatically: offline distillation from execution traces
\citep{ni2026trace2skill,wang2026skillx}, and online evolution through
failure-driven reflection \citep{alzubi2026evoskillautomatedskilldiscovery},
validation-gated updates \citep{yang2026skilloptexecutivestrategyselfevolving},
adversarial co-evolution against a verifier \citep{zhang2026coevoskills},
shared asset layers \citep{ma2026skillclaw}, credit signals
\citep{tu2026dynamic}, reinforcement learning over skill libraries
\citep{xia2026skillrl}, and skill--tool co-evolution
\citep{wei2026skillsmith}; reflective textual evolution can even
outperform scalar-reward RL in sample efficiency \citep{agrawal2025gepa}.
Despite their diversity, these systems share one commitment: acceptance
is decided by a \emph{verifiable} signal---exact match, an executable
check, or a programmatic scorer. Our work targets that shared assumption
and asks whether a judge can stand in for the verifier, and when.

\paragraph{Judges already inside the evaluation loop.}
The assumption is not that judges are absent from this setting, but that
where they appear they are \emph{reference-based}. SealQA
\citep{pham2025sealqa}, a benchmark used throughout the skill-evolution
literature, is scored by a frozen judge model that receives the question,
the gold answer, and the agent's response before returning a binary
verdict. Supplying the gold answer converts judging into a comparison
task and sidesteps the question we study. Our concern is the
\emph{reference-free} regime that a judge gate necessarily occupies:
during optimization no gold answer exists for the candidate being
evaluated, so the judge must supply the standard itself.

\paragraph{Reward-free and judge-based optimization.}
In RL post-training, LLM judges supply learning signal where verifiable
rewards are unavailable, e.g.\ by ranking rollouts \citep{corbitt2025ruler} or
combining preference, judge, and programmatic signals to curb reward
hacking \citep{peng2025agentic}. This work operates in weight space and
assumes the judge provides usable signal; it does not characterize when
that assumption holds. We transpose the judge-signal idea into
frozen-weight, text-space skill optimization and treat judge
trustworthiness as the object of study.

\paragraph{LLM-as-judge reliability.}
LLM judges exhibit systematic biases and can be gamed
\citep{zheng2023judging}. We add a task-dependent axis: a judge's ability to
\emph{evaluate} answers is bounded by its ability to \emph{solve} them,
and this bound---not generic bias---explains where judge gating fails.


\section{Method}
\begin{figure*}[t]
\centering
\includegraphics[width=\textwidth]{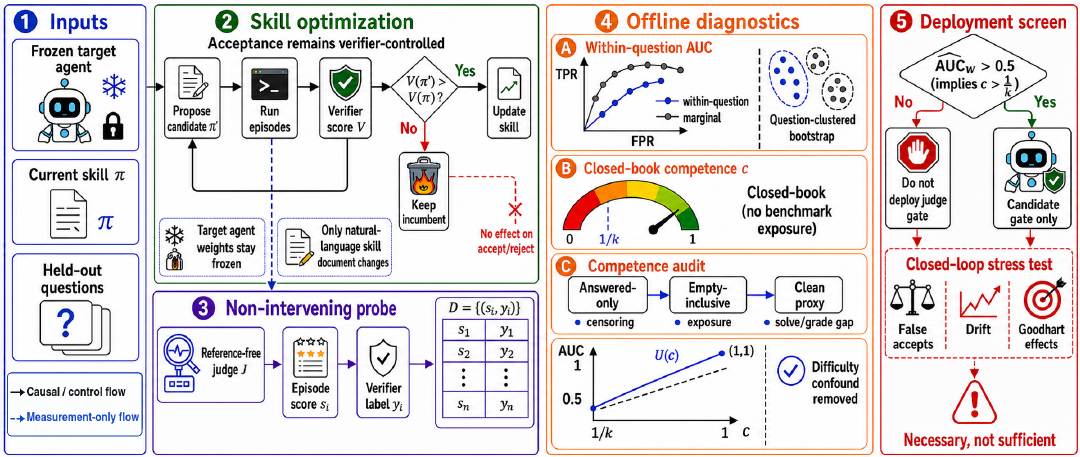}
\caption{\textbf{Overview.}
Skill optimization (2) remains verifier-controlled: $\pi'$ is accepted
only when its verifier score improves, while target weights stay frozen.
A non-intervening probe (3) records $(s_i,y_i)$ per episode without
affecting acceptance, yielding $\mathcal{D}=\{(s_i,y_i)\}$.
Offline diagnostics (4) estimate within-question AUC with a
question-clustered bootstrap, closed-book competence, and the effects of
censoring, exposure, and the solve/grade gap. The deployment screen (5)
rules out judge gating unless competence exceeds the chance floor---a
necessary but not sufficient condition.}
\label{fig:pipeline}
\vspace{-10pt}
\end{figure*}

Figure~\ref{fig:pipeline} gives the overall pipeline. Where does a
judge's verdict get its information, if it cannot see the gold answer? Our starting point is a minimal hypothesis: a
reference-free judge has no source of ground truth independent of its
own ability to solve, so evaluation degenerates informationally into an
implicit re-solving---\emph{to grade, one must first be able to answer}.
We first set up the \emph{judge-as-latent-solver} model, then derive the
closed-form relation between ROC-AUC and competence together with its
testable predictions, and finally present the corresponding measurement
instruments---the non-intervening probe, episode-level estimation, and
competence measurement.


\subsection{Setup and Model Assumptions}
\label{sec:model}
Consider a task whose answer space has size $k \ge 2$ (four-option MCQ
gives $k{=}4$; free-text is the limit $k \to \infty$), with gold answer
$a^{*}$. An \emph{episode} consists of a candidate answer $a$ produced by
the optimized system and its verifier label
$y = \mathbb{1}[a = a^{*}] \in \{0,1\}$. Without access to $a^{*}$, the
judge emits a score $s$ for the pair (question, $a$). Writing $S_1, S_0$
for scores sampled under $y{=}1$ and $y{=}0$, discriminability is
$\mathrm{AUC} := \Pr(S_1 > S_0) + \tfrac12 \Pr(S_1 = S_0)$, the
Mann--Whitney probability that a random positive--negative pair is
correctly ordered.

We model a reference-free judge with three assumptions.
\textbf{(A1) Implicit re-solving:} the judge forms an internal solution
$\hat{a}$, conditionally independent of $a$, with
$\Pr(\hat{a} = a^{*}) = c$, where $c \in [0,1]$ is its \emph{competence
in the grading context}. \textbf{(A2) Error dispersion:} on
$\{\hat{a} \neq a^{*}\}$, $\hat{a}$ is uniform over the remaining $k-1$
answers. \textbf{(A3) Consistency scoring:} $s = \mathbb{1}[a = \hat{a}]$.
The model is in the tradition of Dawid--Skene noisy-annotator models
\citep{dawid1979maximum}, differing in that the annotator's confusion structure
is realized as an explicit re-solving process, which makes $c$
separately measurable by closed-book self-solve accuracy.

\paragraph{Scope of the model.}
(A1) does not assert that verification is universally as hard as
generation. Where a rubric or external evidence supplies a
solution-independent standard, or where a cheap local check suffices---%
units, a boundary condition, one step of a supplied proof---verification
is easier and (A1) understates the judge. We accordingly scope the model
to \emph{single-answer tasks judged without rubric or external evidence},
where the verdict rests on agreement between the candidate and what the
judge would itself conclude. Within that scope (A1) is falsifiable: it
predicts collapse to chance as $c \to 1/k$, which our experiments test.
Rubric-based and evidence-grounded judging are natural extensions.

\subsection{Analytic Characterization of Discriminability}
\label{sec:char}
\noindent\textbf{Proposition 1.} \emph{Under (A1)--(A3),}
\begin{equation}
\mathrm{AUC} = \frac12\left(1 + c - \frac{1-c}{k-1}\right)
             = \frac12 + \frac{ck-1}{2(k-1)}.
\label{eq:prop1}
\end{equation}

\noindent\emph{Proof sketch.} When $y{=}1$, $a = a^{*}$, so
$\mathrm{TPR} = \Pr(\hat{a} = a^{*}) = c$. When $y{=}0$, the event
$\{\hat{a} = a\}$ implies $\{\hat{a} \neq a^{*}\}$, so by (A1)--(A2)
$\mathrm{FPR} = (1-c)/(k-1)$. For a binary score,
$\mathrm{AUC} = \tfrac12(1 + \mathrm{TPR} - \mathrm{FPR})$; substituting
gives \eqref{eq:prop1}. $\square$

Equation~\eqref{eq:prop1} exhibits a threshold structure:
$\mathrm{AUC} - \tfrac12$ has the same sign as $ck - 1$. Two results show
this threshold is robust to realistic departures, while the closed-form
value should be read as an \emph{upper bound}.

\noindent\textbf{Lemma 1 (noise shrinks toward chance).} \emph{Let
$s' = s + \varepsilon$ with $\varepsilon$ i.i.d., independent of
$(s,y)$, continuously distributed. Then
$\mathrm{AUC}' - \tfrac12 = (\mathrm{AUC} - \tfrac12)(2q-1)$ with
$q := \Pr(Z<1) \in (\tfrac12, 1]$, $Z := \varepsilon_0 - \varepsilon_1$.}
Noise therefore shrinks AUC toward $\tfrac12$ by a factor in $(0,1]$
without changing its sign relative to $\tfrac12$.

\noindent\textbf{Remark 1 (collusion and upper-boundedness).} Relaxing
(A1)--(A2) to specify only the \emph{collusion rate}
$\rho := \Pr(\hat{a} = a \mid y{=}0)$ gives
$\mathrm{AUC} = \tfrac12(1 + c - \rho)$, strictly decreasing in $\rho$.
(A1)--(A2) correspond to $\rho = (1-c)/(k-1)$. When judge and evaluated
model err in correlated ways---both drawn to the same distractor, typical
of models sharing a training distribution---$\rho$ exceeds that value, so
at fixed $c$ Proposition~1 is an attainable upper bound. With Lemma~1,
the model's commitment is to \emph{bounds and thresholds, not a fitted
curve}, matching the necessary-condition phrasing we keep throughout.

\noindent\textbf{Corollary 1 (chance-floor threshold).}
\emph{(i) Under (A1)--(A3), with or without the noise of Lemma~1,
$\mathrm{AUC} > \tfrac12 \Longleftrightarrow c > 1/k$. (ii) In the
relaxed family of Remark~1 with super-uniform collusion
$\rho \ge (1-c)/(k-1)$, only the forward implication survives:}
\begin{equation}
\mathrm{AUC} > \tfrac12 \;\Longrightarrow\; c > 1/k .
\label{eq:cor1}
\end{equation}

\noindent\emph{Proof.} (i) is immediate from \eqref{eq:prop1}, whose sign
relative to $\tfrac12$ is that of $ck-1$; Lemma~1 rescales the deviation
by a strictly positive factor and cannot change its sign. For (ii),
Remark~1 gives $\mathrm{AUC} > \tfrac12 \iff c > \rho$; combining with
$\rho \ge (1-c)/(k-1)$ yields $c(k-1) > 1-c$, i.e.\ $c > 1/k$. $\square$

The converse fails in the relaxed family: whenever $\rho \ge c$---judge
and candidate erring together at least as often as the judge is
right---discriminability is at or below chance even though $c > 1/k$.
\textbf{We therefore state the chance-floor threshold throughout as a
\emph{necessary but not sufficient} condition}: above-chance competence
must hold for a judge gate to discriminate at all, but it does not
guarantee that it does. Correspondingly $\mathrm{AUC} \downarrow \tfrac12$
as $c \downarrow 1/k$, and low-$k$ tasks whose models share a training
distribution are precisely where $\rho$ is expected to be large.

\noindent\textbf{Corollary 2 (free-text limit).}
$\lim_{k\to\infty} \mathrm{AUC} = (1+c)/2$. Two independent errors almost
never coincide verbatim in free text ($\rho \to 0$), so errors cannot
collude and a unit of competence buys the most discriminability; at small
$k$ the judge's wrong solution may coincide with---and thereby
endorse---the candidate's wrong answer. Discriminability is thus ordered
jointly by $(c,k)$, not by task difficulty alone.

\noindent\textbf{Corollary 3 (AUC-inversion competence meter).}
Under (A1)--(A3), $c$ is identified by $(\mathrm{AUC}, k)$:
$\hat{c}_{\mathrm{AUC}} = [(k-1)(2\,\mathrm{AUC}-1) + 1]/k$, degenerating
to $2\,\mathrm{AUC}-1$ in the free-text limit. By Lemma~1 and Remark~1
both noise and super-uniform collusion depress the observed AUC, so
$\hat{c}_{\mathrm{AUC}}$ is systematically \emph{conservative}.
Methodologically this matters because it is constructed purely from
grading behavior, never routing through the closed-book answering path,
and is therefore an estimator orthogonal to benchmark contamination.

\noindent\textbf{Remark 2 (difficulty heterogeneity confounds the
marginal AUC).} Let item $t$ have target accuracy $p_t$ and judge
competence $c_t$, with (A1)--(A3) holding conditionally. Tilting the item
distribution by class yields
\begin{equation*}
\begin{aligned}
\mathrm{AUC}_{\mathrm{marg}}
&= \underbrace{\tfrac12\Big(1+\bar{c}-\tfrac{1-\bar{c}}{k-1}\Big)}_{\text{Prop.~1 at }\bar{c}}
\;+\; \tfrac12\,\mathrm{Cov}(p_T,c_T)\,\Delta_k, \\[2pt]
\Delta_k &:= \tfrac{1}{\mathbb{E}[p_T]} - \tfrac{1}{(k-1)\,\mathbb{E}[1-p_T]}.
\end{aligned}
\end{equation*}
When difficulty acts on both models, $\mathrm{Cov}(p_T,c_T) > 0$:
positive episodes are enriched in easy items, negatives in hard ones. If
$\mathbb{E}[p_T] < (k-1)/k$---always true in the free-text limit---then
$\Delta_k > 0$ and the marginal AUC is systematically inflated.
In the extreme, a judge that scores only by perceived difficulty and never
reads the answer attains $\mathrm{AUC} > \tfrac12$ with \emph{no}
answer-level discrimination. The marginal AUC therefore does not identify
the quantity of interest.

\noindent\textbf{Proposition 2 (stratified discriminability is
invariant).} \emph{Define the within-question AUC
$\mathrm{AUC}_w := \sum_t \omega_t \mathrm{AUC}_t$, pairing positives and
negatives only within the same item, with $\omega_t$ the normalized count
of such pairs. Then (i) Proposition~1 holds per item; (ii) by linearity in
$c_t$, $\mathrm{AUC}_w$ equals \eqref{eq:prop1} at the pair-weighted
competence $\tilde{c} := \sum_t \omega_t c_t$; and (iii) $\mathrm{AUC}_w$
is invariant to any additive item-only score component $g(T)$, since all
episodes in a stratum shift by the same constant.}

\noindent\textbf{Corollary 4 (stratified threshold).}
$\mathrm{AUC}_w > \tfrac12 \iff \tilde{c} > 1/k$: the necessary condition
survives difficulty heterogeneity, with the competence parameter refined
from the population mean to the pair-weighted mean. The gap
$\mathrm{AUC}_{\mathrm{marg}} - \mathrm{AUC}_w$ directly estimates the
difficulty-confound share, and we report both side by side.

\subsection{The Non-Intervening Probe}
\label{sec:probe}
Let the optimizer \citep{yang2026skilloptexecutivestrategyselfevolving} hold the incumbent skill $\pi$ and
accept a candidate $\pi'$ by the rule
$\mathbb{1}[V(\pi') > V(\pi)]$ on a verifier score $V$. The quantity in
Proposition~1 requires the pairs $(s_i, y_i)$ to be drawn from the
episode distribution induced by the \emph{genuine} optimization process:
once judge scores enter the acceptance rule, selection pressure changes
the candidate distribution, and what one measures is discriminability
already distorted by that pressure.

We therefore record judge scores $J(\cdot)$ on the selection set in
parallel while leaving the acceptance rule unchanged. Since the decision
is a deterministic function of $V$ and $J$ does not enter its domain, the
probed and unprobed runs are path-wise identical in distribution; the
probe's entire output is the episode-level sample
$\mathcal{D} = \{(s_i,y_i)\}_{i=1}^{n}$. The instrument therefore
measures discriminability on the distribution induced by the
\emph{verifier} gate---a \emph{necessary} condition for a usable judge
gate; sufficiency under judge control is a separate question we take up in
the closed-loop study below.

\subsection{Estimating Episode-Level Discriminability}
\label{sec:est}
Given $\mathcal{D}$ with positive and negative index sets
$\mathcal{P}, \mathcal{N}$, we estimate discriminability by the
Mann--Whitney statistic
$\widehat{\mathrm{AUC}} = \frac{1}{n_1 n_0}\sum_{i \in \mathcal{P}}
\sum_{j \in \mathcal{N}} [\mathbb{1}(s_i > s_j) + \tfrac12
\mathbb{1}(s_i = s_j)]$, and its stratified counterpart
$\widehat{\mathrm{AUC}}_w$ by restricting pairs to a common item and
weighting strata by their pair counts; only items carrying both classes
form valid strata. We report both, their difference estimating the
difficulty-confound share. Because episodes cluster within items,
confidence intervals for both come from a \emph{question-clustered
bootstrap}---episode-level resampling would understate variance. AUC is
chosen over accuracy because it is exactly the quantity characterized by
Propositions~1--2, and because it is insensitive to class imbalance and to
strictly monotone rescaling of scores.

Two design choices follow directly from the model. \emph{Granularity:}
the consistency event $\mathbb{1}[a = \hat{a}]$ is defined per episode;
averaging scores at the skill level erases this structure before
measurement and leaves only sparse samples, since candidate skills update
only occasionally. Episode-level scoring builds $\mathcal{D}$ from stored
rollouts with no reruns. \emph{Answer-content feed:} (A3) requires $s$ to
be a function of the answer's \emph{content}. A feed that renders $s$
independent of $a$---for instance a bare multiple-choice letter with no
option text---implies $s \perp y$ and $\mathrm{AUC} \equiv \tfrac12$
under the model: a measurement artifact unrelated to judge competence. An
early version of our probe fell into exactly this degenerate case and
produced a spurious null. We feed answer content and report both the
chosen option's text and the full reasoning; the model further predicts a
small advantage for the latter, as intermediate derivations give the
judge's implicit re-solving checkable alignment points.

\subsection{Measuring Competence and Decontamination}
\label{sec:measure}

\paragraph{Closed-book self-solve.}
The judge answers each task without access to the gold answer, which is used only for offline scoring. We adopt the \emph{empty-inclusive} accuracy
$\hat{c}_{\mathrm{solve}}=\frac1n\sum_t\mathbb{1}[\hat a_t=a_t^*]$,
counting episodes that fail to produce a parseable answer within the generation budget as unsolved. Answered-only accuracy would introduce survivorship bias, since models are more likely to fail to converge on items they cannot solve.

\paragraph{Two notions of competence.}
The parameter $c$ in (A1) denotes grading-context competence,
$c_{\mathrm{grade}}$, whereas closed-book evaluation measures
$c_{\mathrm{solve}}$. Because the candidate answer and its reasoning may scaffold re-solving, we expect
$c_{\mathrm{solve}}\le c_{\mathrm{grade}}$ in recall-friendly domains, making $c_{\mathrm{solve}}$ a lower-bound proxy. Since
$U(c)=\tfrac12+(ck-1)/(2(k-1))$ increases with $c$, an observed AUC below
$U(c_{\mathrm{solve}})$ conservatively supports the bound. An observation above it does not falsify the theory, as it may reflect either difficulty confounding or underestimation of $c_{\mathrm{grade}}$. Likewise, we invoke the chance-floor test only when a plausible upward correction still leaves $c_{\mathrm{solve}}$ near $1/k$.

\paragraph{Cross-validation under contamination.}
Accuracy on a public benchmark $B$ may be inflated by memorized question--answer mappings, which may transfer differently between direct answering and rollout grading. We therefore use two estimators with distinct error sources. First, we measure
$\hat c_{\mathrm{solve}}(B')$ on a domain- and difficulty-matched but less-exposed benchmark $B'$, with
$\widehat{\Delta}=\hat c_{\mathrm{solve}}(B)-\hat c_{\mathrm{solve}}(B')$
estimating memorization inflation. Second, we use the grading-only inversion estimate
$\hat c_{\mathrm{AUC}}$. The former depends on the cleanliness and match of $B'$, whereas the latter depends on the model assumptions and is conservative. If the two agree and both lie well below the surface accuracy, then the conclusion that the headline figure is inflated does not rely on either estimator alone.

\section{Experiments}
\label{sec:exp}
\subsection{Setup}
The optimizer is SkillOpt, the target Haiku, the judge Claude Sonnet
(fixed throughout), with its prompt reusing the optimizer's template. We
evaluate three tasks spanning distinct competence regimes: research
mathematics (MCQ), factual QA (free-text), and GPQA-Diamond
\citep{rein2023gpqa} (graduate-science MCQ), with episode counts
(pos/neg) 210 (87/123), 360 (223/137), and 198 (141/57).

\subsection{Discriminability Varies Sharply Across Tasks}
Table~\ref{tab:disc} reports discriminability for the primary judge
(Claude Sonnet). The marginal AUC ranges from near-random on mathematics
(0.46) to strong on factual QA (0.86), with GPQA in between (0.74). By
Remark~2, however, the marginal estimate conflates answer-level
discrimination with a shared difficulty axis; the identifying quantity is
the within-question estimate $\widehat{\mathrm{AUC}}_w$
(Proposition~2), which pairs only positive and negative episodes of the
\emph{same} question. On factual QA a large difficulty-confound share
($+0.12$) is removed, dropping the honest reading to $0.735$; on
mathematics the two agree ($\approx 0$). GPQA is evaluated in a single
pass and admits no within-question strata, so only the (optimistic)
marginal AUC is available. Even after stratification, the task ordering is
preserved and mathematics stays near chance: judge signal is
task-dependent and cannot be assumed.

\begin{table}[t]
\centering
\small
\setlength{\aboverulesep}{0pt}\setlength{\belowrulesep}{0pt}
\setlength{\tabcolsep}{4pt}
\begin{tabular}{lcccc}
\toprule
Task & $k$ & $\widehat{\mathrm{AUC}}$ & $\widehat{\mathrm{AUC}}_w$ & confound \\
\midrule
Research math & 5 & 0.457 & 0.489 & $-0.03$ \\
Factual QA    & $\infty$ & 0.855 & 0.735 & $+0.12$ \\
GPQA-Diamond  & 4 & 0.735 & n/a & --- \\
\bottomrule
\end{tabular}
\caption{Marginal vs.\ within-question discriminability (Sonnet judge),
with the difficulty-confound share $\widehat{\mathrm{AUC}}-
\widehat{\mathrm{AUC}}_w$. 0.5 is chance. GPQA has no within-question
strata.}
\label{tab:disc}
\end{table}

\subsection{Testing the Closed-Form Bound}
\label{sec:bound}
Proposition~1 predicts an \emph{upper bound} on discriminability given an
independent measurement of competence. We supply that independent
measurement with closed-book self-solve accuracy
$\hat{c}_{\mathrm{solve}}$; note that
$\hat{c}_{\mathrm{AUC}}$ cannot serve here, since inversion and forward
prediction are mutual inverses and the test would be vacuous.
Table~\ref{tab:bound} reports the comparison for two judges. The
prediction to be tested is the inequality, not equality: by Lemma~1 and
Remark~1, label-independent noise and super-uniform collusion both
depress the observed value below the bound.

\begin{table}[t]
\centering
\small
\setlength{\aboverulesep}{0pt}\setlength{\belowrulesep}{0pt}
\setlength{\tabcolsep}{4pt}
\begin{tabular}{llccccc}
\toprule
Task & Judge & $k$ & $\hat{c}_{\mathrm{solve}}$ & pred. & obs. & obs $\le$ pred \\
\midrule
Math     & Sonnet & 5 & 0.267 & 0.542 & 0.489$^{w}$ & \checkmark \\
Factual  & Sonnet & $\infty$ & 0.567 & 0.783 & 0.735$^{w}$ & \checkmark \\
Factual  & Haiku  & $\infty$ & 0.633 & 0.817 & 0.634$^{w}$ & \checkmark \\
GPQA     & Sonnet & 4 & 0.626 & 0.751 & 0.735$^{m}$ & \checkmark \\
GPQA     & Haiku  & 4 & 0.540 & 0.694 & 0.794$^{m}$ & $\times$ \\
\bottomrule
\end{tabular}
\caption{Closed-form bound (Proposition~1) against observation.
$^{w}$within-question $\widehat{\mathrm{AUC}}_w$; $^{m}$marginal
$\widehat{\mathrm{AUC}}$ (GPQA admits no strata). The single violation
falls in the only cell lacking both a stratified estimate and a
contamination-free competence measurement; see text.}
\label{tab:bound}
\vspace{-10pt}
\end{table}

The bound holds on every task that admits within-question
stratification. The one violation, GPQA$\times$Haiku, falls in the only
cell meeting neither condition for a clean test, and two mechanisms are
jointly consistent with its $+0.10$ excess. First, GPQA is evaluated in a
single pass, so no strata exist and the observation is necessarily the
\emph{marginal} AUC, which Remark~2 shows is inflated by the difficulty
confound; the sign of the excess matches that prediction. Second,
$\hat{c}_{\mathrm{solve}}$ lower-bounds grading-context competence,
so $U(\hat{c}_{\mathrm{solve}}) = 0.694$ may be
evaluated below $U(c_{\mathrm{grade}})$: a grading competence of $0.66$,
well within the gap our estimators span here, would place the bound above
the observation. Separating the two requires a task that is both
contamination-controlled and multiply evaluated.

Turning to the chance floor, mathematics is the clearest case for
Corollary~1: $\hat{c}_{\mathrm{solve}} = 0.267$ against
$1/k = 0.2$, and even a substantial scaffolding correction leaves
competence near the floor, where discriminability should be barely above
$\tfrac12$. We observe $0.489$, marginally below chance. Competence this
close to the floor cannot support a usable gate, while competence above
it does not guarantee one.

\subsection{Decomposing the Gap Between Benchmark and Genuine Competence}
\begin{figure}[t]
\centering
\includegraphics[width=0.98\columnwidth]{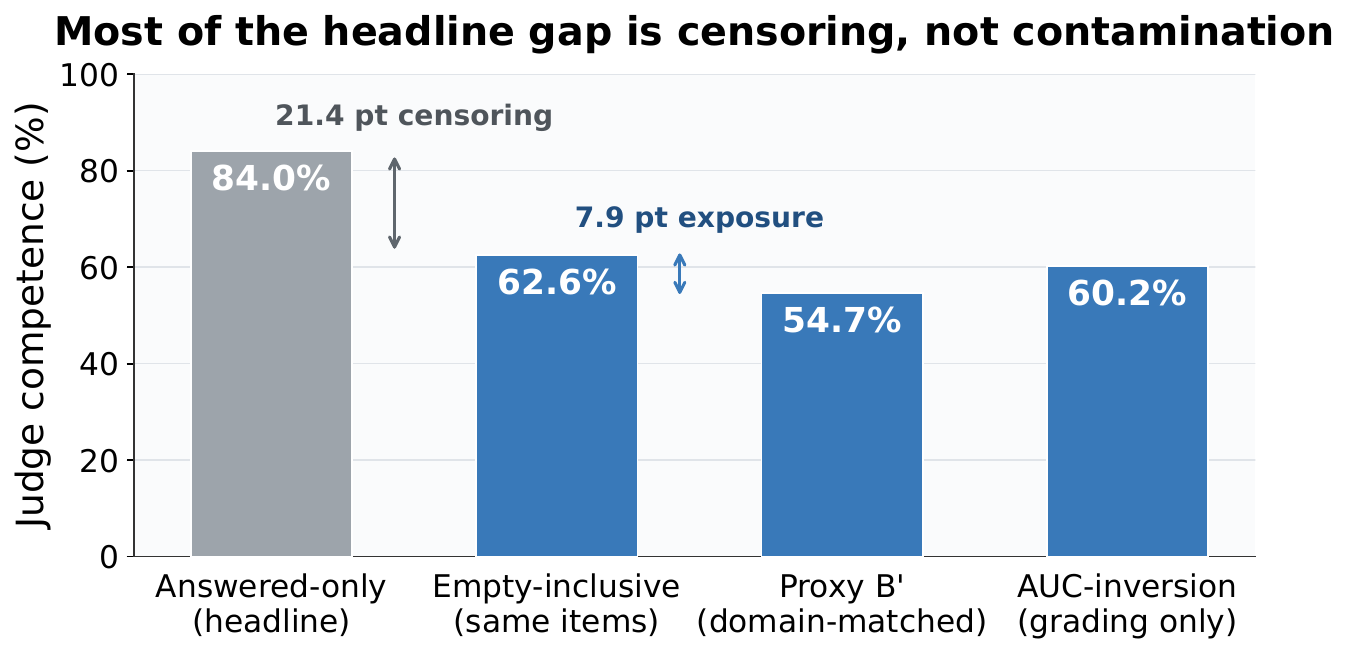}
\caption{Competence-gap decomposition for Sonnet on GPQA-Diamond.}
\label{fig:contam}
\vspace{-10pt}
\end{figure}

Corollary~3 supplies an estimator built purely from grading behavior,
bypassing the answering path where memorized question--answer mappings
are cued. Table~\ref{tab:three} places it beside the proxy estimate and
two surface figures. The headline gap for Sonnet---$0.840$ reported against $\approx 0.55$
genuine (Figure~\ref{fig:contam})---decomposes into distinct mechanisms that we separate rather than attribute wholesale to contamination.

\begin{table}[t]
\centering
\small
\setlength{\aboverulesep}{0pt}\setlength{\belowrulesep}{0pt}
\setlength{\tabcolsep}{4.5pt}
\begin{tabular}{lcccc}
\toprule
Judge & Answered-only & Empty-incl. & Proxy $B'$ & Inversion \\
\midrule
Sonnet & 0.840 & 0.626 & 0.547 & 0.602 \\
Haiku  & ---   & 0.540 & 0.353 & 0.691 \\
\bottomrule
\end{tabular}
\caption{Competence estimates on GPQA-Diamond. The first two differ only
in counting non-convergent episodes; the last two are independent
estimators with orthogonal errors.}
\label{tab:three}
\vspace{-10pt}
\end{table}

\paragraph{(i) Censoring, not exposure: $0.840 \to 0.626$.}
The larger part of the gap, $21.4$ points, is a scoring convention. Both
figures are the same judge on the same items; they differ only in whether
episodes that fail to converge within the generation budget enter the
denominator. Reasoning judges fail to converge precisely on items they
cannot solve, so discarding them selects for solvable items. This is
survivorship bias and has nothing to do with benchmark exposure; it is,
however, the convention under which headline accuracies are frequently
reported, which is why we flag it first.

\paragraph{(ii) Exposure: $0.626 \to 0.547$.}
The part plausibly attributable to benchmark exposure
\citep{golchin2024time} is the residual $7.9$ points
between the empty-inclusive GPQA figure and a domain-matched proxy $B'$:
the hard-difficulty physics, chemistry and biology subset of SuperGPQA
\citep{du2026supergpqa}, same subjects as GPQA-Diamond but larger and
more recent, hence less per-item exposure. This is what our
decontamination design estimates, and it is materially smaller than the
raw gap. It also inherits the proxy's uncertainty: $B'$ matches on domain
and difficulty but is not the same instrument, so part of these $7.9$
points may reflect dataset mismatch rather than memorization.


\paragraph{(iii) The inversion estimate, and estimator disagreement.}
For Sonnet, the inversion estimate corroborates the other measures:
$\hat{c}_{\mathrm{AUC}}=0.602$ lies between the empty-inclusive and proxy
estimates. Despite different error sources---scoring convention, dataset
match, and model assumptions---all three fall in the $0.55$--$0.63$ range,
well below the answered-only headline. For Haiku, however, the proxy
($0.353$) and inversion ($0.691$) differ by $0.34$, so we do not claim
agreement. As noted above, the inversion estimates
$c_{\mathrm{grade}}$, whereas the proxy measures $c_{\mathrm{solve}}$.
Their gap may therefore reflect a stronger scaffolding effect for the
weaker judge, which benefits more from seeing a candidate answer. Thus,
the two estimates may bracket Haiku's competence, but too widely to
support a point estimate. \textbf{We therefore restrict the agreement
claim to Sonnet and treat Haiku as a case of estimator divergence}, also
consistent with the competence-mismatch explanation for its bound
violation above. For deployment, the key point is
directional: censoring, exposure, and the solve/grade gap all make the
uncorrected benchmark figure an optimistic input to the chance-floor
test.

\subsection{Cross-Provider Generalization}
\begin{figure}[t]
\centering
\includegraphics[width=\columnwidth]{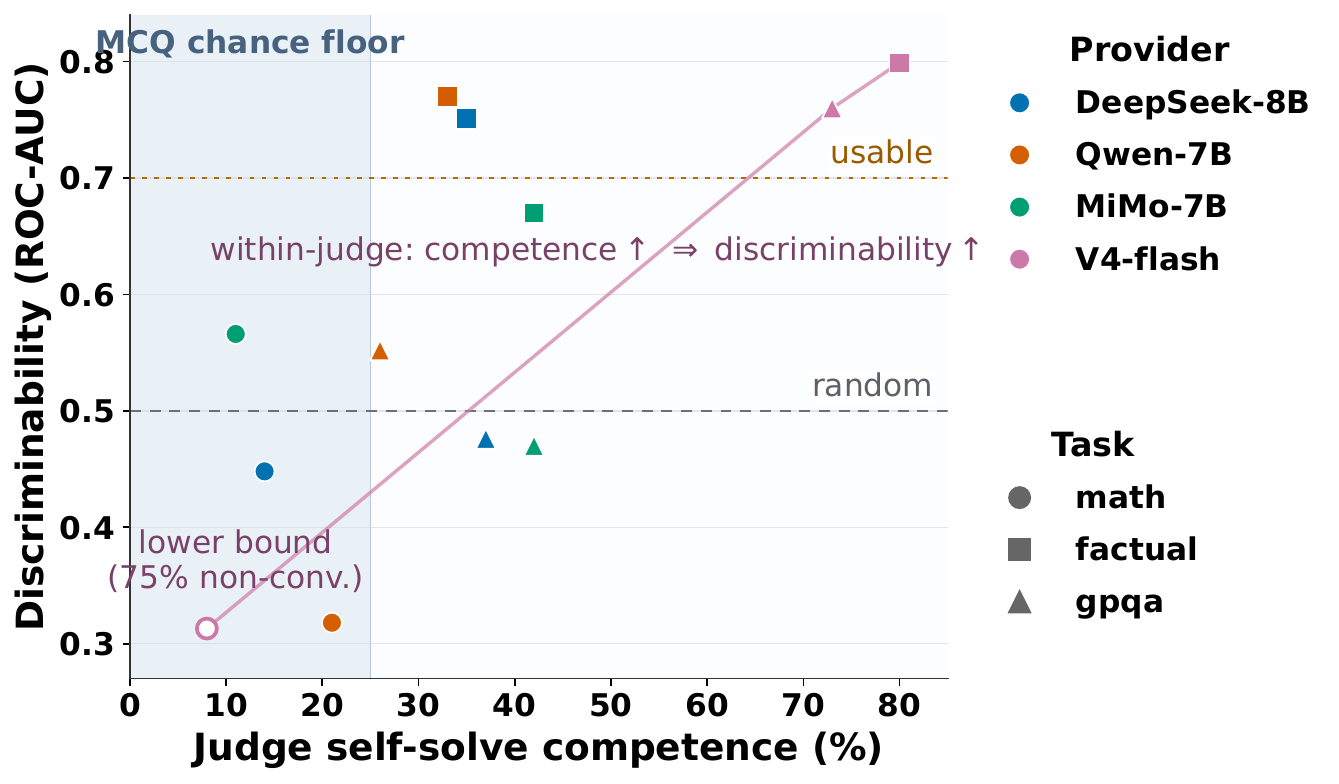}
\caption{Discriminability versus genuine competence across four judges and three
tasks. It approaches chance near the competence floor and becomes usable
only clearly above it. The solid line connects V4-flash across
tasks, giving within-judge evidence.}
\label{fig:main}
\vspace{-10pt}
\end{figure}

To test whether the competence--discriminability relationship generalizes,
we repeat the diagnostic with four judges from independent model
families: DeepSeek-8B, Qwen-7B, Xiaomi MiMo-7B, and DeepSeek-V4-flash (a
reasoning model), through the identical harness path with a matched
16k-token budget. Table~\ref{tab:cross} and Figure~\ref{fig:main} report
empty-inclusive self-solve competence (non-convergent episodes counted as
unsolved) and discriminability.

\begin{table}[t]
\centering
\small
\setlength{\aboverulesep}{0pt}\setlength{\belowrulesep}{0pt}
\setlength{\tabcolsep}{4pt}
\begin{tabular}{lcccccc}
\toprule
& \multicolumn{2}{c}{Math} & \multicolumn{2}{c}{Factual} & \multicolumn{2}{c}{GPQA} \\
\cmidrule(lr){2-3}\cmidrule(lr){4-5}\cmidrule(lr){6-7}
Judge & comp. & AUC & comp. & AUC & comp. & AUC \\
\midrule
DeepSeek-8B & 14\% & 0.45 & 35\% & 0.75 & 37\% & 0.48 \\
Qwen-7B     & 21\% & 0.32 & 33\% & 0.77 & 26\% & 0.55 \\
MiMo-7B     & 11\% & 0.57 & 42\% & 0.67 & 42\% & 0.47 \\
V4-flash    & 8\%$^{*}$ & 0.31 & 80\% & 0.80 & 73\% & 0.76 \\
\bottomrule
\end{tabular}
\caption{Cross-provider competence (empty-inclusive) and
discriminability. $^{*}$high non-convergence (75\% empty); competence is
a bound.}
\label{tab:cross}
\vspace{-10pt}
\end{table}

The task-level pattern reproduces across every provider: on mathematics,
where competence is at or below the chance floor (8--21\%),
discriminability is near-random (0.31--0.57); on factual QA, where
competence is moderate-to-high (33--80\%), it is usable (0.67--0.80).
That four independently trained families reproduce the same
collapse-on-math / usable-on-factual pattern indicates the relationship
is not an artifact of any single provider. DeepSeek-V4-flash is the
clearest case because it varies competence \emph{within a single judge}:
competent on GPQA (73\%) and discriminable (0.76), yet not competent on
mathematics (8\%) and near-random (0.31)---holding the judge fixed and
giving within-judge evidence consistent with competence as an important
driver.

\paragraph{Measurement note.}
Reasoning judges emit long chains of thought and frequently exhaust the
generation budget without emitting an answer. We count such
non-convergent episodes as unsolved rather than excluding them: an
answered-only accuracy inflates competence by survivorship (V4-flash
scores 94.8\% over answered GPQA items but 73.0\% once non-convergence is
included). We flag V4-flash on mathematics (75\% non-convergence) as a
bound, consistent with genuine inability. This matrix covers the four
open-weight judges, and so replicates on independent families the pattern
first observed on Claude.

\paragraph{Estimator caveat.}
This matrix reports marginal AUC, since per-episode scores were not
retained for these judges. The confound it carries lands where it does
not bear on the claim: in our main analysis the share is $+0.12$ to
$+0.19$ on free-text factual QA but $\approx 0$ on mathematics---the
column that carries the collapse result.

\subsection{From Discriminability to Closed-Loop Safety}
\label{sec:closedloop}
The probe deliberately withholds control from the judge, which raises the
question of whether the offline screen anticipates how a judge behaves
once it controls acceptance. We test this at small scale. For a fixed
judge we measure $\widehat{\mathrm{AUC}}_w$ on stored rollouts, then run
optimization under gates that differ only in the signal driving
accept/reject: the verifier score, the judge score, or a random gate. In
the judge and random arms the verifier is used only for offline
evaluation and never enters a decision. Target and optimizer models are
fixed; only the gate varies.

\begin{table}[t]
\centering
\small
\setlength{\aboverulesep}{0pt}\setlength{\belowrulesep}{0pt}
\setlength{\tabcolsep}{3.5pt}
\begin{tabular}{llcccc}
\toprule
Task & Gate & $\widehat{\mathrm{AUC}}_w$ & Final & FA & FR \\
\midrule
Factual & Verifier       & ---   & 0.908 & 0.00 & 0.00 \\
        & Judge (passes) & 0.735 & 0.825 & \textbf{0.00} & 0.44 \\
        & Judge (fails)  & 0.620 & 0.842 & 0.33 & 0.33 \\
        & Random         & ---   & 0.875 & 0.33 & 0.50 \\
\midrule
Math    & Verifier       & ---   & 0.363 & 0.00 & 0.00 \\
        & Judge (fails)  & 0.425 & 0.238 & 0.50 & 0.50 \\
        & Random         & ---   & 0.350 & 0.25 & 0.25 \\
\bottomrule
\end{tabular}
\caption{Diagnostic discriminability against closed-loop outcome.
``Final'' is the held-out verifier score of the final skill, averaged
over seeds (3 for factual QA, 2 for math); FA/FR are false-accept and
false-reject rates. ``Passes'' denotes a judge whose
$\widehat{\mathrm{AUC}}_w$ interval excludes $\tfrac12$.}
\label{tab:loop}
\vspace{-10pt}
\end{table}

Table~\ref{tab:loop} separates two questions the aggregate score
conflates: whether gating on a judge \emph{helps}, and what mistakes it
makes---the latter being where the screen is informative.

\paragraph{Failing the screen predicts harmful acceptance.}
Both judges whose discriminability interval includes or falls below
$\tfrac12$ accept candidates that measurably hurt: false-accept rates of
$0.33$ on factual QA and $0.50$ on mathematics, the latter matching a
coin flip. On mathematics, where the point estimate is \emph{below}
$\tfrac12$---a weakly anti-correlated signal---the judge gate is worse
than random ($0.238$ vs.\ $0.350$; per-seed ranges $[0.225, 0.250]$ and
$[0.325, 0.375]$ do not overlap). A judge that fails the screen is thus
not merely uninformative but can be actively harmful, since optimizing
against an anti-correlated signal drives the system away from
improvement rather than leaving it where it started.

\paragraph{Passing the screen avoids harmful acceptance in this pilot.}
The judge that passes behaves qualitatively differently: its false-accept
rate is $0.00$---across seeds it never admitted a candidate that hurt---%
and it issued the same number of accepts and rejects as the verifier
gate. Its errors are one-sided over-rejection ($\mathrm{FR} = 0.44$): it
forgoes some improvements rather than importing regressions, and on the
aggregate score it is conservative, matching the random gate rather than
the verifier ($0.825$ vs.\ $0.875$ vs.\ $0.908$; the judge and random
per-seed ranges $[0.800, 0.850]$ and $[0.825, 0.925]$ overlap). At this
scale, then, the screen is informative about the \emph{composition} of a
gate's errors---and specifically about harmful acceptance---rather than
about final performance, which depends on further factors it does not
capture.

\paragraph{Interpretation.}
Read as a screen, clearing the chance floor is a \emph{prerequisite} for
safe judge-driven optimization, not a guarantee of competitive
performance---predicting the latter needs criteria beyond
discriminability. These are preliminary observations: at two to three
seeds they support ordering and error-type claims, not effect-size
estimates, and on mathematics the verifier gate itself barely separates
from random ($0.363$ vs.\ $0.350$), so the informative comparison there
is judge against random.

\section{Conclusion}
Before an LLM judge can gate skill optimization, one must know whether its
scores carry signal at all. Modeling a reference-free judge as a latent
solver makes this precise: discriminability is bounded by the judge's
competence, giving a necessary---though not sufficient---condition,
competence above the chance floor. A non-intervening probe
provides support on the tested tasks---discriminability approaches chance
where measured competence is near the floor, and a judge that fails the
screen can gate worse than randomly. The
upshot is a cheap pre-deployment screen: estimate uncontaminated
competence on the target task, and decline to gate on a judge that does
not clear the floor. The competence that governs discriminability is the
judge's \emph{genuine} ability, not its headline benchmark accuracy---a
distinction that matters most exactly where contamination is most likely,
on the standard benchmarks a practitioner would reach for first. Whether
a judge that clears the screen survives sustained optimization---under
drift and Goodhart pressure---is the question our necessary condition
leaves open, and the natural next step.

\bibliography{references}

\end{document}